\documentclass[10pt,twocolumn,letterpaper]{article}
\usepackage{multirow}
\usepackage{cvpr}
\usepackage{times}
\usepackage{epsfig}
\usepackage{graphicx}
\usepackage{amsmath}
\usepackage{amssymb}
\usepackage{colortbl}
\usepackage{booktabs} 

\usepackage{lipsum}

\usepackage[breaklinks=true,bookmarks=false]{hyperref}

\cvprfinalcopy 

\def\cvprPaperID{****} 

\begin{document}

\title{Attacking Convolutional Neural Network using Differential Evolution}

\author{
Jiawei Su\\
Kyushu University\\
Japan\\
{\tt\small jiawei.su@inf.kyushu-u.ac.jp}
\and
Danilo Vasconcellos Vargas\\
Kyushu University\\
Japan\\
{\tt\small vargas@inf.kyushu-u.ac.jp}
\and
Kouichi Sakurai\\
Kyushu University\\
Japan\\
{\tt\small sakurai@csce.kyushu-u.ac.jp}
}

\maketitle

\begin{abstract}
   The output of Convolutional Neural Networks (CNN) has been shown to be discontinuous which can make the CNN image classifier vulnerable to small well-tuned artificial perturbations. That is, images modified by adding such perturbations(i.e. adversarial perturbations) that make little difference to human eyes, can completely alter the CNN classification results.
In this paper, we propose a practical attack using differential evolution(DE) for generating effective adversarial perturbations. We comprehensively evaluate the effectiveness of different types of DEs for conducting the attack on different network structures. The proposed method is a black-box attack which only requires the miracle feedback of the target CNN systems.
The results show that under strict constraints which simultaneously control the number of pixels changed and overall perturbation strength, attacking can achieve $72.29\%$, $78.24\%$ and $61.28\%$ non-targeted attack success rates, with $88.68\%$, $99.85\%$ and $73.07\%$ confidence on average, on three common types of CNNs. The attack only requires modifying 5 pixels with 20.44, 14.76 and 22.98 pixel values distortion.
Thus, the result shows that the current DNNs are also vulnerable to such simpler black-box attacks even under very limited attack conditions.
\end{abstract}

\section{Introduction}

\begin{figure}[t]
\begin{center}
\includegraphics[width=0.8\linewidth]{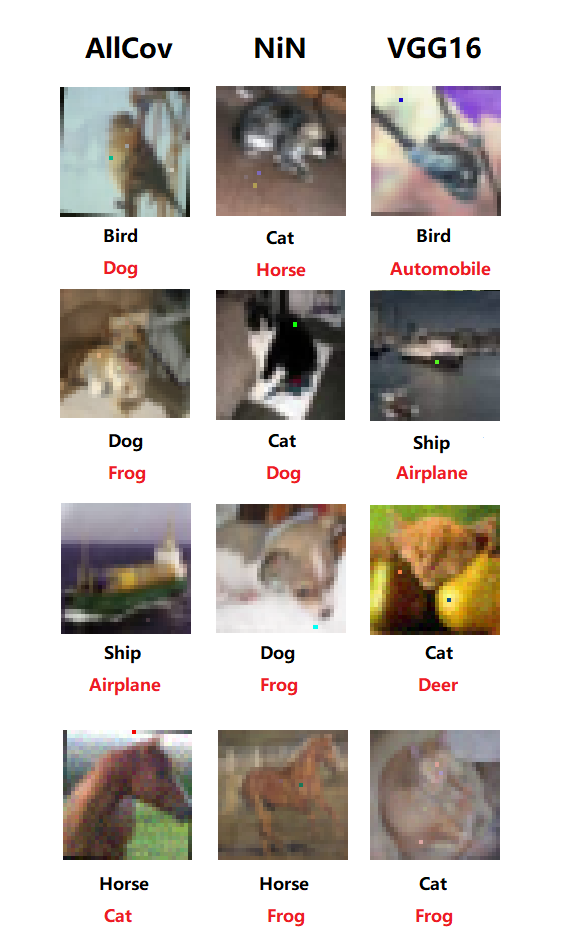}
\end{center}
   \caption{The proposed few-pixel attack that successfully fooled three types of DNNs trained on CIFAR-10 dataset: The All convolutional network(AllConv), Network in network(NiN) and VGG. The original class labels are in black color while the target class labels are in blue.}
\label{our_results}
\label{fig:long}
\label{fig:onecol} 
\end{figure}

Recent research has shown that Deep Convolutional Neural Network(CNN) can achieve human-competitive accuracy on various image recognition tasks \cite {9}. 
However, several recent studies have suggested that the mapping learned by CNN from input image data to the output classification results, is not continuous. That is, there are some specific data points (or possibly some continuous regions) in the input space whose classification labels can be changed by adding even very small perturbations. Such modification is called ``adversarial perturbation'' in the case that potential adversaries wish to abuse such a characteristic of CNN to make it misclassify\cite {1, 2, 3, 4}. 
By using various optimization methods, tiny well-tuned additive perturbations which are expected to be imperceptible to human eyes but be able to alter the classification results drastically, can be calculated effectively. In specific, adding the adversarial perturbation can lead the target CNN classifier to either a specific or arbitrary class, both are different from the true class.

In this research, we propose and evaluate a black-box method of generating adversarial perturbation based on differential evolution, a natural inspired method which makes no assumptions about the problem being optimized and can effectively search very wide area of solution space.
Our proposal has mainly the following contribution and advantages compared to previous works:
 
\begin{itemize}

\item \textbf{Effectiveness} - With the best parameter setting of DE and extremely limited conditions, the attack can achieve $72.29\%$, $78.24\%$ and $61.28\%$ success rates of conducting non-targeted attacks on three types of common convolutional neural network structures: Network in Network, All convolutional Network and VGG16\cite{32} trained on CIFAR-10 dataset. Further results on ImageNet dataset show that non-targeted attacking the BVLC AlexNet model can alter the labels of $31.87\%$ of the validation images. 


\item \textbf{Black-Box Attack} - The proposed attack only needs miracle reaction (probability labels) from the target CNN system while many previous attacks requires access to the inner information such as gradients, network structures, training data and so on, which in most cases is hard or even not available in practice.  
The capability of being able to conduct black-box attack using DE is based on the fact that it makes no assumption on the optimization problem of finding effective perturbation such that does not abstract the problem to any explicit target functions according to the assumption, but works directly on increasing(decreasing) the probability label values of the target(true) classes.

\item \textbf{Efficiency} - Many previous attacks of creating adversarial perturbation require alternation on a considerable amount of pixels such that it may risk the possibility being perceptible to human recognition systems as well as require higher cost of conducting the modification (i.e. the more pixels that need to be modified, the higher the cost). The proposed attack only requires modification on 5 pixels with an average distortion of 19.39 pixel value per channel. Specifically, the modification on 5 pixels is further pressured by adding a term that is proportional to the strength of accumulated modification in the fitness functions of DEs.

\item \textbf{Scalability} - Being able to attack more types of CNNs (e.g. networks that are not differentiable or when the gradient calculation is difficult) as long as the feedback of the target systems is available.
\end{itemize}

The rest of the paper is as follows: Section~2 introduces previous attack methods and their features, as well as compares with the proposed method. Section~3 describes why and how to use DE to generate effective adversarial perturbation under various settings. In Section~4, several measures are proposed for evaluating the effectiveness of DE based attack. Section~5 discusses the experimental results and points out possible future extension.
\begin{figure}[t]
\begin{center}
\includegraphics[width=0.7\linewidth]{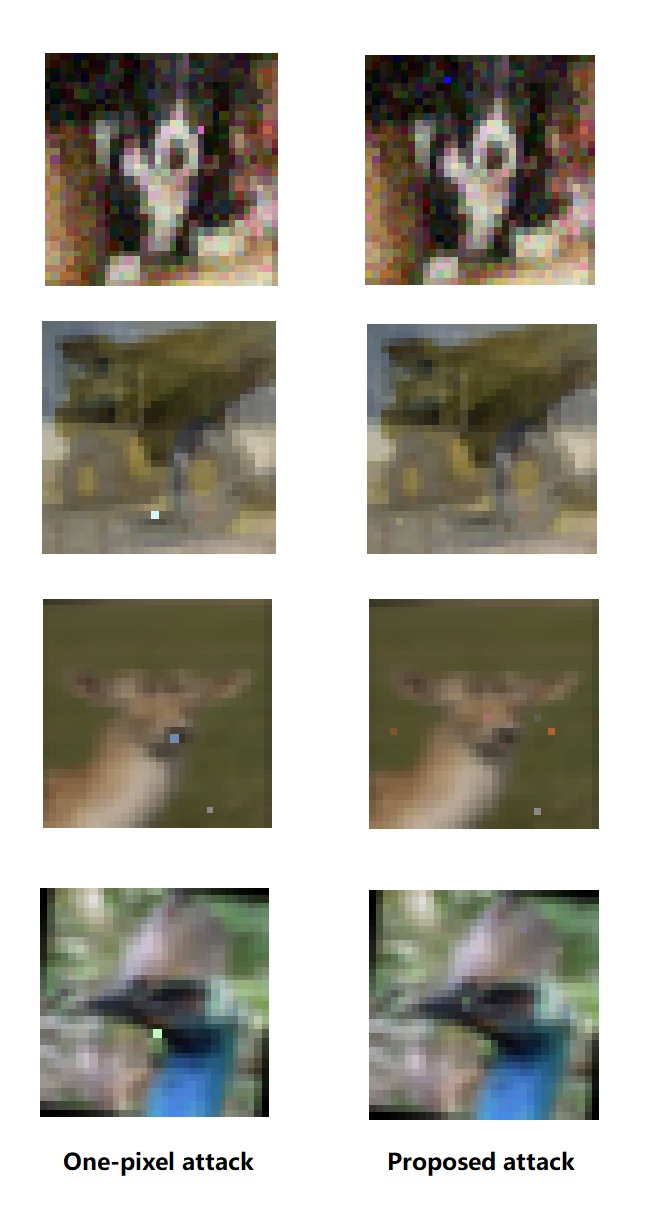}
\end{center}
   \caption{A comparison between the adversarial images generated by proposed attack and one-pixel attack. Since the former has control mechanisms embedded in the fitness function, the distortion it caused is expected to be less perceptible than one-pixel attack. As can be seen, even if requiring perturbing more pixels, the proposed attack can have similar or better visual effect to one-pixel attack in practice which only few or even none of the perturbed pixels are noticeable.}
\label{fig:long}
\label{fig:aaa}
\label{fig:onecol} 
\end{figure}



\section{Related works}

Though CNN has given outstanding performance of classification in different practical domains, its security problem has been also emphasized\cite {6} \cite{barreno2010security}. For example, in the domain of natural language processing, the CNN-based text classification can be easily fooled by purposely adding or replacing specific words or letters \cite{40}. For speech-to-text recognition, the signal can be also altered by adding a tiny additional signal such that the resulting text can be very different from the origin \cite{42}. 
The CNN-based image recognition suffers the same problem. In fact, the intriguing(or vulnerable) characteristic that CNN is sensitive to well-tuned artificial perturbation was first reported by evaluating the continuity of CNN output with respect to small change on input image \cite {3}. Accordingly various optimization approaches are utilized for generating effective perturbation to attack the CNN image classifiers. 
I.J.Goodfellow et al. proposed ``fast gradient sign'' algorithm for calculating effective perturbation based on a hypothesis in which the linearity and high-dimensions of inputs are the main reason that a broad class of networks are sensitive to small perturbation \cite {2}. 
S.M. Moosavi-Dezfooli et al. proposed a greedy perturbation searching method by assuming the linearity of CNN decision boundaries \cite {4}. 
N. Papernot et al. utilize Jacobian matrix with respect to the network to build ``Adversarial Saliency Map'' which indicates the effectiveness of conducting a fixed length perturbation through the direction of each axis \cite {1, 20}. Based on these preliminary works, attacks in extreme conditions are also proposed to show the vulnerability of CNN is even more serious. One pixel attack shows that one bad pixel can be able to alter the entire classification output \cite{41}. Unlike common image-specific perturbations, the universal adversarial perturbation is a single constant perturbation that can fool a large amount of images at the same time \cite{28}.

To the best of our knowledge, the one-pixel attack is the only existing work which implemented DE for finding optimized adversarial perturbation\cite{41}. The work shows that DE can generate effective solution even under very limited condition(i.e. only one-pixel can be modified). However, the one-pixel attack only aims to show the possibility of conducting the attack with DE and implements one kind of simple DE with a constant F value as 0.5 and no crossover, which leaves the problem of evaluating and comparing other kinds of different DE variants. 
The proposed few-pixel attack indeed modifies more pixels than one-pixel attack. However, it does not mean that few-pixel attack requires more access to the target image since even for the one-pixel attack, it is also necessary to access to all pixels of the image to find the best pixel to perturb.
In addition, one-pixel attack does not fully consider the constraints in practice, for example there is no terms for controlling the distortion of pixels in the fitness function used by one-pixel attack.
On the other side, the proposed few-pixel attack still requires modification on less pixels compared to most previous works.
Furthermore, in this research we focus on non-targeted attacks while one-pixel attack is based on targeted-attack. Due to the significant difference on successful rate of attack and other factors such as time and resource consumption, the simpler non-targeted attack can be more practical, especially in the case of large-scale attack. A comparison between the proposed and one-pixel attack showing the difference on methodologies is summarized by Figure.~3 (See Section~4 for the detail description).
Other black-box attacks that require no internal knowledge about the target systems such as gradients, have also been proposed. 
N. Papernot et al. proposed the first black-box attack against CNN which consists in training a local model to substitute for the target CNN, using inputs synthetically generated by an adversary and labeled by the target CNN. The local duplication is then used for crafting adversarial examples which are found being successfully misclassified by the targeted CNN \cite{25}. N. Narodytska et al, implemented a greedy local-search to perturb a small set of pixels of an image which treats the target CNN as a miracle \cite{24}. 
\begin{figure}[t]
\begin{center}
\includegraphics[width=0.9\linewidth]{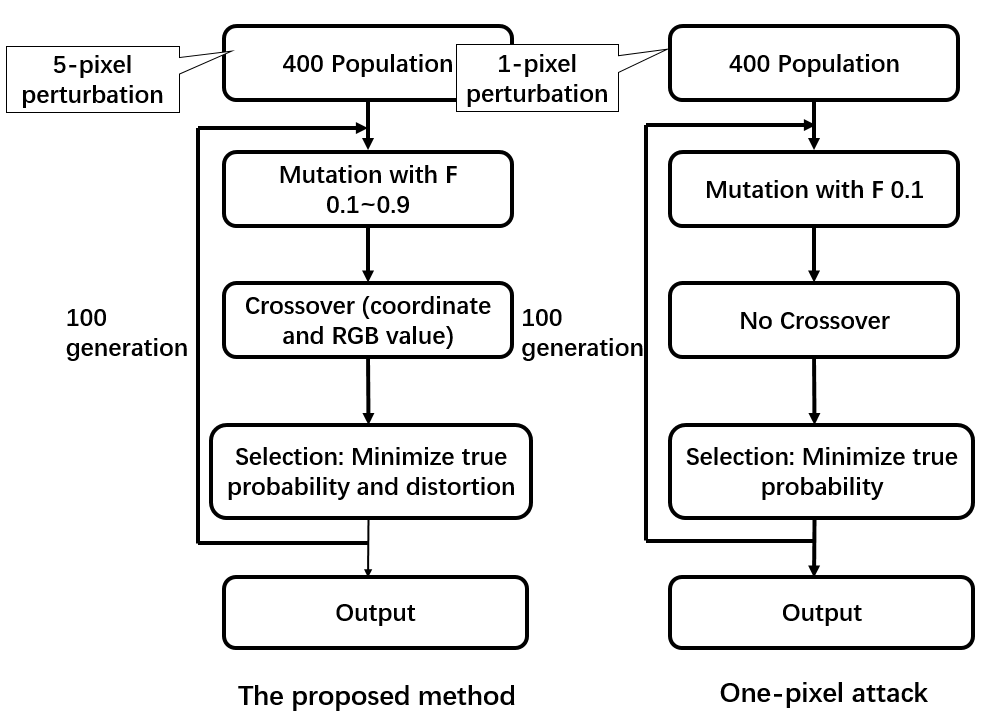}
\end{center}
   \caption{A comparison between the proposed attack and one-pixel attack showing the difference on parameter settings.}
\label{fig:long}
\label{fig:aaa}
\label{fig:onecol} 
\end{figure}
\section{Methodology}
\subsection{Problem Description}

Calculating adversarial perturbation added to a natural image for confusing CNN classification can be abstracted as an optimization problem with constraints. 
Assuming that a 2-d three-channel RGB image can be represented by a flattened n-dimensional vector in which each scalar element represents a tuple consisting of three channel values of a pixel.
Let $f$ be the target image classifier which receives n-dimensional inputs, $\textbf{x} = (x_1,..,x_n)$ be the original natural image classified with predicted label $c_p$ according to $f$. Note that $c_p$ might not be the ground true of $\textbf{x}$ since $f$ can also misclassify without any outside interfering. 
The soft label (i.e. probability) of sample $\textbf{x}$ being with label $c_p$ is represented as $f_{c_p}(\textbf{x})$. 
A vector $e(\textbf{x}) = (e_1,..,e_n)$ which has the same dimensions to $\textbf{x}$ represents a specific additive perturbation with respect to a specific natural image $\textbf{x}$, which is being able to alter the label of $\textbf{x}$ from $c_p$ to the target class $t_{adv}$ where $c_p \neq t_{adv}$ with the modification strength less than maximum modification limitation $L$, which for example can be measured by the length of vector $e(\textbf{x})$ (e.g. the accumulated pixel values modified) or the number of none-zero elements of $e(\textbf{x})$ (i.e. the number of pixels modified). 
Therefore the ultimate goal of adversaries is to find the optimized solution $e(\textbf{x})^{*}$ for the following question. In the case of targeted attack, the target class $t_{adv}$ is designated while for non-targeted attack, it can be an arbitrary class as long as $t_{adv} \neq  c_p$.

\begin{equation*}
\begin{aligned}
& \underset{e(\textbf{x})^{*}}{\text{maximize}}
& & f_{t_{adv}}(\textbf{x}+e(\textbf{x})) \\
& \text{subject to}
& & \Vert e(\textbf{x}) \Vert \leq L
\end{aligned}
\end{equation*}

In the case of this research, the maximum modification limitation $L$ is set to be two empirical constraints: 1) The number of pixels can be modified, which is represented by $d$, is set to be 5 while the specific index of each modified pixel is not fixed. The constraint can be represented as $\Vert e(\textbf{x}) \Vert_{0} \leq d$ where $d = 5$. Except the elements need to modify, others in vector $e(\textbf{x})$ are left to zero. 2) The fitness functions of DE utilized in this research favor the modification with smaller accumulated pixel values more than success rate of attack such that controlling the accumulated pixel values becomes the priority during the evolution. Such constraints are more restricted compared to many previous works which only implement restrctions similar to either 1) or 2) \cite{28, 29}.



Geometrically, the entire input space of a CNN image classifier can be seen as a n-dimensional cube such tat the proposed attacks that modifies 5 pixels are essentially searching the effective perturbation on the 5-dimensional slices of input space where $5 \leq d$, which the size of the slice is further limited by constraints on distortion implemented in the fitness function. In other words, the search of each iteration of DE is limited to towards 5 directions also with limited length of steps which each direction is perpendicular to a certain axes of the input space coordinate. 
However, the probe can still logically find an arbitrary data-point in the input space by using multiple iterations. Even if in each iteration, the search directions and area of the proposed attack are limited, it can still probe the entire input space towards arbitrary 3-d direction to find better optima by iterating the progress. This is illustrated in Figure~ for the case when $n=3$ and $d=2$. 

\begin{figure}[t]
\begin{center}
\includegraphics[width=0.7\linewidth]{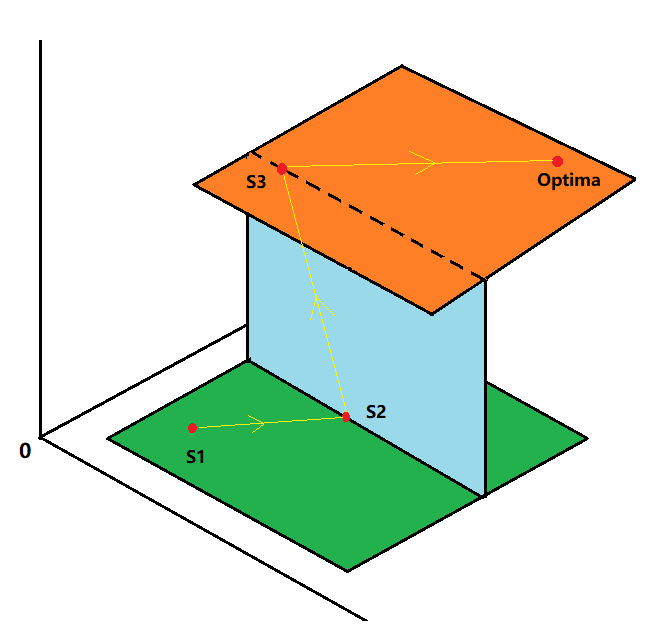}
\end{center}
   \caption{An illustration of conducting two-pixel perturbation attack in a 3-dimensional input space coordinate(i.e. the image has three pixels). The original natural image is a data point represented by S1. Due to the limitation on the number of dimensions that can be probed, in each iteration the search is only allowed on a 2-d plane (shown by green, blue and orange planes in the figure) around the current solution. As shown, after three iterations which the direction of probe is shown by yellow arrows, it finds the optimal point. By iterating the evolution of DE, the 2-d probe can actually move in towards arbitrary directions in 3-d space to find the optima.}
\label{fig:long}
\label{fig:onecol} 
\end{figure}

\subsection{Perturbation Strength}
In this research, a five-pixel modification is choose as the strength of attack by considering the practicability of the attack. First, few-pixel modification is more efficiency than global perturbation \cite{28} that modifies each or most pixels of an image due to less variables need to solve. On the other side, one-pixel attack numerically requires the least cost \cite{41}. However, the one-pixel attack can be hard imperceptible in practice since all attack strength concentrates on the single modified pixel. By adding the number of pixels that can modify, the strength can be distributed to make the modification less visible. In practice, a scenario that one-pixel attack is available but five-pixel attack is not common. A visual comparison of the proposed five-pixel attack and one-pixel attack is illustrated by Figure.~2.

\subsection{Differential Evolution and Its Variants}

Differential evolution (DE) is currently one of the most effective stochastic real-parameter optimization method for solving complex multi-modal optimization problems \cite{12}, \cite{das2011differential}. Similar to Genetic algorithms and other evolutionary algorithms, DE acts as a black-box probe which does not care the specific form of the target functions. Thus, it can be utilized on a wider range of optimization problems (e.g, non-differentiable, dynamic, noisy, among others). DE uses iterative progress for improving the quality of the population which each individual in the population, which is always called genome, is a potential solution for the corresponding target problem. In particular, DE considers difference of the individual genomes as search ranges within each iteration to explore the solution space. In addition, DE uses one-to-one selection holds only between an ancestor and its offspring which is generated through mutation and recombination, rather than the commonly used tournament selection in many other evolutionary algorithms. Such a selection strategy has a superior ability to preserve population diversity better than tournament selection where elites and their offspring may dominate the population after few iterations. \cite{civicioglu2013conceptual}.

Different DE variants mainly demarcate from others on the ways of conducting mutation and crossover. We specifically introduce how to combine different strategies of mutation and crossover for implementing various kinds of DEs. The specific settings of DEs implemented in this research are summarized by Table~4 and Table~5.

\subsubsection{Mutation}
In the biological point of view as well as genetic algorithms inspired, the mutation is a random change on an individual $x_i$ of the population in order to gain higher diversity through iterations. Being different from genetic algorithms, which directly conduct random change of values on $x_i$, one of the most basic mutation strategies is to randomly choose three other individuals, indexed by $r1, r2$ and $r3$ from the current population, and combine them with a scale parameter $F$ to form the mutated $x_i$, denoted by ${x_i}^*$. Specifically, the ${x_i}^*$ is obtained from the following formula:

\begin{eqnarray*}
& {x_i}^* = x_{r1} + F(x_{r2} + x_{r3}),\\
& r1\not = r2\not = r3,
\end{eqnarray*}

where $F$ is the scale parameter set to be in the range from 0 to 1. It can be seen that under such a scheme, the mutated ${x_i}^*$ has no relationship with its prototype ${x_i}$. Their relations can be established in the crossover step. 

The intuition of such a mutation is using the individual $x_{r1}$ as the basis, plus the difference (scaled by the factor $F$) between another two individuals $x_{r2}$ and $x_{r3}$ to generate child. Such difference indicates a meaningful step in the search space. It is actually the different values of parameter $F$ demarcates from one mutation to another. Instead of a constant $F$, it can be also set to be random and to be specific for each individual in a certain iteration. In this research, we respectively adopt different $F$ to evaluate the influence to the attack success rates.

\subsubsection{Crossover}

The crossover step after mutation, concerns about combining the original individual ${x_i}$ and its corresponding child ${x_i}^*$. This is the step that ${x_i}$ and ${x_i}^*$ actually establish the connection to each other, which is used for improving the potential diversity of the population. Specifically, the crossover exchanges the components of ${x_i}^*$ obtained by mutation step, with the corresponding elements of its prototype ${x_i}$, by using two kinds of crossover strategies: exponential crossover and binomial crossover. 

Simply put, the exponential crossover replaces a series of elements of ${x_i}^*$, saying any elements without the range from index $i$ to $j$, with the elements of ${x_i}$ that own the same index, where $1 \leqslant i \leqslant j \leqslant D$ where $D$ is the size of an individual. 
On the other hand, binomial crossover replaces every element of ${x_i}^*$ according to a probability of crossover, denoted by $C_r$. Specifically, a random number within the range from 0 to 1 is generated for each element in ${x_i}^*$, replace with the corresponding value of ${x_i}$ if it is smaller than $C_r$.

Each individual (genome) of DE holds the information of one five-pixel attack (perturbation). That is, each individual represents a series of perturbation on five pixels, which the information of each pixel perturbation includes its x-y coordinate position and RGB value. Hence an individual is encoded in a 5X5 array. 

Simply put, one single perturbation consists of its location of conducting perturbation and specific values of perturbation. We consider an approach by combining exponential and binomial crossover such that the new crossovers probabilistically exchange these two types of information between a currently individual and its offspring. Specifically, we consider the following 4 types of crossovers:

\begin{itemize}

\item \textbf{Crossover on position information.}
The crossover only replaces the position information (i.e. the first two dimensions) of ${x_i}^*$ with the one owned by ${x_i}$. A probability value $C_p$ is used to identify if the crossover triggers or not. 
Exchanging information of coordinate is for letting the offspring inherits the location information of vulnerable pixels containing in current population.

\item \textbf{Crossover on RGB values.}
The crossover only replaces the RGB value information (i.e. the last three dimensions) of ${x_i}^*$ with the one owned by ${x_i}$. A probability value $C_{rgb}$ is used to identify if the crossover triggers or not. 
Exchanging information of coordinate is for letting the offspring inherits the information of vulnerable RGB perturbation values containing in current population.  

\item \textbf{Crossover for both position and RGB values.}
Such a crossover is the combination of the above two, according to the assumption that both crossovers are useful.

\item \textbf{No crossover.}
The opposite to the one above, assuming that exchanging either information of pixel locations or RGB values is not meaningful.

\end{itemize}
	
\subsubsection{Selection}

The selection step implemented by this research makes no difference to the standard DE selection setting. Specifically, unlike the tournament selection in Genetic Algorithms which ranks all population based on the individual fitness and selects amount of best individuals, DE uses a one-to-one selection holds only competitions between a current individual ${x_i}$ and its offspring ${x_i^*}$ which is generated through mutation and crossover. This ensures that DE retains the very best so-far solution at each index therefore the diversity can be well preserved.

\subsubsection{Other DE variants}
It is worth to mention that even if different variants of DE have been implemented and evaluated in this research, there are actually even more complex variations/improvements such as self-adaptive \cite{brest2006self},  multi-objective \cite{vargas2015general}, among others, which can potentially further improve the effectiveness of attack.

\subsection{Using Differential Evolution for Generating Adversarial Perturbation}

The use of DE for generating adversarial images have the following main advantages:

\begin{itemize}

\item \textbf{Higher probability of Finding Global Optima} - DE is a meta-heuristic which is relatively less subject to local minima than gradient descent or greedy search algorithms (this is in part due to diversity keeping mechanisms and the use of a set of candidate solutions). Capability of finding better solutions (e.g. global optima rather than local) is necessary in our case since we have implemented more restricted constraints on perturbation in this research such that the quality of optimization solution has to be guaranteed to a high extent.

\item \textbf{Require Less Information from Target System} -  DE does not require the optimization problem to be differentiable as is required by classical optimization methods such as gradient descent and quasi-newton methods. 
This is critical in the case of generating adversarial images since 1) There are networks that are not differentiable, for instance \cite {5}.  
2) Calculating gradient requires much more information about the target system which can be hardly realistic in many cases. 

\item \textbf{Simplicity} - The approach proposed here is independent of the classifier used. For the attack to take place it is sufficient to know the probability labels. In addition, most of previous works abstract the problem of searching the effective perturbation to specific optimization problem (e.g. an explicit target function with constraints). Namely additional assumptions are made to the searching problem and this might induce additional complexity. Using DE does not solve any explicit target functions but directly works with the probability label value of the target classes.

\end{itemize}

\subsection{Method and Settings}

The DE is used to find the best perturbation which can achieve high probability label of target class and low modification strength. The information of a proposed five-pixel perturbation (which is one individual of the DE population) is encoded into an five-dimensional array which each dimension contains five elements: x-y coordinates and RGB value of one-pixel perturbation. 
The initial number of population is $400$ and during each iteration another $400$ candidate solutions (children) will be produced by various types of mutation and crossover. Then a 400 knock-out selection is conducted between each pair of individual and its offspring, to produce the new population with the same size to the last generation. The fitness function used is as follows:

\begin{eqnarray*}
& F{(x_i)} = 0.25P_t(x_i)+0.75C(x_i),\\
& C(x_i) = (R(x_i)+G(x_i)+B(x_i))/256,
\end{eqnarray*}

where $F{(x_i)}$ is the fitness value of an individual $x_i$, which is a combination of its probability value belonging to the true class t, $P_t(x_i)$, and the cost of attack $C(x_i)$. Weight values of 0.25 and 0.75 are respectively assigned to the two terms. We find that a higher weight value assigned to $P_t(x_i)$ will make the DE evolution take much less care of $C(x_i)$ such that the cost of attack increases drastically. While doing the opposite will increase $P_t(x_i)$ but less significantly. Such weights indicate that obtaining a $x_i$ with low $P_t(x_i)$ is much easier than a $x_i$ with low $C(x_i)$. The cost $C(x_i)$ is measured as average pixel value changed on each pixel modified, which is expected to be small to guarantee the modification can be invisible. For an individual, the lower the fitness, the better the quality hence easier the survival.

The maximum number of generation is set to $100$ and early-stop criteria can be triggered when there is at least one individual in the population whose fitness is less than 0.007. 
Once stopped, the label of true class is compared with the highest non-true class to evaluate if the attack succeeded. The initial population is initialized by using uniform distributions $U (1, 32)$ for CIFAR-10 images for generating x-y coordinate (e.g. the image has a size of 32X32 in CIFAR-10) and Gaussian distributions N ($\mu$=128, $\sigma$=127) for RGB values. For ImageNet the setting is similar.

\subsection{Finding the Best Variant}
In order to find the best DE variant for generating adversarial samples, we propose a greedy-search method which starts from a DE variant with basic setting. Then we gradually alter the parameter settings to evaluate the effect on the success rate of attack and come up with a local-optimized setting, which is further used for attack under several different scenarios. Specifically, it is mainly the mutation and crossover that differ different types of DE variants. We implement a basic DE which enables both mutation and crossover to middle levels. Then we adjust the value of each single parameter while keep others unchanged to conduct the test.  

For example, the four types of crossover proposed in Section~3.3.2, can be achieved by adjusting the corresponding crossover probability $C_p$ and $C_{rgb}$. For instance, both $C_p$ and $C_{rgb}$ are set to be a very small number means to disable the crossover.

\section{Evaluation and Results}
The following measures are utilized for evaluating the effectiveness and efficiency of the proposed attack: 

\begin{itemize}

\item \textbf{Success Rate} - It is defined as the empirical probability of a natural image that can be successfully altered to another pre-defined (targeted attack) and arbitrary class (non-targeted attack) by adding the perturbation.

\item \textbf{Confidence} - The measure indicates the average probability label of the target class output from the target system when successfully altered the label of the image from true to target.

\item \textbf{Average distortion}

The average distortion on the single pixel attacked by taking the average modification on the three color channels, is used for evaluating the cost of attack. Specifically, the cost is high if the value of average distortion is high such that it is more likely to be perceptible to human eyes.

\end{itemize}

\subsection{Comparison of DE variants and Further Experiments}
Preliminary experiments are for evaluating different DEs (i.e. different F value and crossover strategies). We utilize a greedy search approach to find the local-optimized DE variant. Specifically, we first propose a standard model which enables all settings to mid levels. Then the settings are gradually changed one-by-one for evaluating the influence to the effectiveness of attack. The local-optimized model is found for conducting further experiments with more datasets and network structures.

Specifically, the comparison of DE variants are conducted on the All convolution network \cite {21} by launching non-targeted attacks against them for finding a local-optimized model. 
The local-optimized model is further evaluated on Network in Network\cite {31} and VGG16 network\cite {32} trained on Cifar-10 dataset \cite {11}. 
At last, the model is applied for non-targeted attacking the BVLC AlexNet network trained on ImageNet dataset with the same DE paramater settings used on the CIFAR-10 dataset, although ImageNet has a search space 50 times larger than CIFAR-10, to evaluate the generalization of the proposed attack to large images.
Given the time constraints, we conduct the experiment without proportionally increasing the number of evaluations, i.e. we keep the same number of evaluations. 

The structures of the networks are described by Table 1, 2 and 3. The network setting were kept as similar as possible to the original with a few modifications in order to achieve the highest classification accuracy. All of them are with ReLu activation functions. For each of the attacks on the three types of Cifar-10 neural networks $500$ natural image samples are randomly selected from the test dataset to conduct the attack. For BVLC AlexNet we use $250$ samples from ILSVRC 2012 validation set selected randomly for the attack. 
 
\begin{table}
\begin{center}
\begin{tabular}{|c|c|}
\hline
conv2d layer(kernel=3, stride = 1, depth=96) \\
conv2d layer(kernel=3, stride = 1, depth=96) \\
conv2d layer(kernel=3, stride = 2, depth=96) \\
conv2d layer(kernel=3, stride = 1, depth=192) \\
conv2d layer(kernel=3, stride = 1, depth=192) \\
dropout(0.3) \\
conv2d layer(kernel=3, stride = 2, depth=192) \\
conv2d layer(kernel=3, stride = 2, depth=192) \\
conv2d layer(kernel=1, stride = 1, depth=192) \\
conv2d layer(kernel=1, stride = 1, depth=10) \\
average pooling layer(kernel=6, stride=1) \\
flatten layer \\
softmax classifier \\
\hline
\end{tabular}
\end{center}
\caption{All convolution network}

\begin{center}
\begin{tabular}{|c|c|}
\hline
conv2d layer(kernel=5, stride = 1, depth=192) \\
conv2d layer(kernel=1, stride = 1, depth=160) \\
conv2d layer(kernel=1, stride = 1, depth=96) \\
max pooling layer(kernel=3, stride=2) \\
dropout(0.5) \\
conv2d layer(kernel=5, stride = 1, depth=192) \\
conv2d layer(kernel=5, stride = 1, depth=192) \\
conv2d layer(kernel=5, stride = 1, depth=192) \\
average pooling layer(kernel=3, stride=2) \\
dropout(0.5) \\
conv2d layer(kernel=3, stride = 1, depth=192) \\
conv2d layer(kernel=1, stride = 1, depth=192) \\
conv2d layer(kernel=1, stride = 1, depth=10) \\
flatten layer \\
softmax classifier \\
\hline
\end{tabular}
\end{center}
\caption{Network in Network}

\begin{center}
\begin{tabular}{|c|c|}
\hline
conv2d layer(kernel=3, stride = 1, depth=64) \\
conv2d layer(kernel=3, stride = 1, depth=64) \\
max pooling layer(kernel=2, stride=2) \\
conv2d layer(kernel=3, stride = 1, depth=128) \\
conv2d layer(kernel=3, stride = 1, depth=128) \\
max pooling layer(kernel=2, stride=2) \\
conv2d layer(kernel=3, stride = 1, depth=256) \\
conv2d layer(kernel=3, stride = 1, depth=256) \\
conv2d layer(kernel=3, stride = 1, depth=256) \\
max pooling layer(kernel=2, stride=2) \\
conv2d layer(kernel=3, stride = 1, depth=512) \\
conv2d layer(kernel=3, stride = 1, depth=512) \\
conv2d layer(kernel=3, stride = 1, depth=512) \\
max pooling layer(kernel=2, stride=2) \\
conv2d layer(kernel=3, stride = 1, depth=512) \\
conv2d layer(kernel=3, stride = 1, depth=512) \\
conv2d layer(kernel=3, stride = 1, depth=512) \\
max pooling layer(kernel=2, stride=2) \\
flatten layer \\
fully connected(size=2048) \\
fully connected(size=2048) \\
softmax classifier \\
\hline
\end{tabular}
\end{center}
\caption{VGG16 network}

\end{table}

\subsection{Results}
The success rates, confidence and perturbation strength for the attack using different DE variants on All convolutional network is shown by Table~4 and Table~5. Then local-optimized DEs are selected to conduct further experiments on three additional types of networks: Network in Network (NIN), VGG16 network and AlexNet BVLC network. The first two are trained on Cifar-10 dataset and the latter is based on ImageNet dataset. The results are shown by Table~6.

Each type of DE variant is abbreviated in the format ``$F value$/$C_p$/$C_{rgb}$''. For example, 0.5/0.5/0.5 denotes the model with its F value, crossover rate of coordinate and RGB value all equal to 0.5. We choose the 0.5/0.5/0.5 as the standard prototype model to compare with other variants, since it enables all settings to a mid extent.

\subsubsection{Effectiveness and Efficiency of Attack.}

First the influence of changing F value is evaluated by implementing the standard model with different F values. According to the results of first 4 rows in Table~4, higher F values give very limited increase on success rate of attack however require a considerable amount of more distortion. For example, shifting from 0.1/0.5/0.5 to 0.9/0.5/0.5 increases only $1.37\%$ success rate with a cost of increasing 5.39($26.53\%$) pixel value. Since the F controls how far the distance starting from the current individuals to probe new solutions, the intuition of this result indicates that moving smaller steps in the solution space might find new solutions that are similar to the prototypes, with comparative attack success rate but more efficient, while moving larger steps may find totally different solutions with higher distortion required. This might indicate that in the solution space, the candidate solutions (vulnerable pixels) are gathered within several groups and moving by small steps from the existing solutions can find new individuals with better quality(i.e. require less distortion). Therefore it comes to a conclusion that smaller F values can effectively decrease the distortion needed for the attack.

Then we keep the F value as 0.5 for conducting further experiments of comparing influence of two crossover strategies. The results show that generally both types of crossover are not helpful for improving success rate and decreasing distortion required. For example, comparing 0.5/0.1/0.1 which disables both crossovers, and 0.5/0.1/0.9(0.5/0.9/0.1) which only enables one crossover, shows $1.25\%$($0.23\%$) reduction on success rate and only 0.04(0.48) decrease on distortion. Enabling both crossovers (0.5/0.9/0.9) is also not helpful in a similar way.
Such results show that the quality of perturbation can not be significantly improved by replacing the coordinate or RGB color information of children population with their corresponding ancestors'.

According to the results of comparison, we choose the 0.5/0.1/0.1 and 0.1/0.1/0.1 as the two local-optimized models for conducting further experiments. Note that as mentioned above, setting a smaller F value can be helpful for decreasing the distortion on perturbed pixels. On CIFAR-10, the success rates of proposed attacks on three types of networks show the generalized effectiveness of the proposed attack through different network structures. 
Specifically, the Network in Network structure shows the greatest vulnerability which gives highest success rate, confidence and least distortion under the same settings of DE. The VGG16 network on the other side, shows the average highest robustness. Attacking the All convolutional network comparatively requires the most distortion and gives mid performance. In addition, a smaller F value is effective for reducing distortion through different network structures.

On ImageNet, the results show that the proposed attack can be generalized to large size images and fool the corresponding larger neural network. 
Note that the ImageNet results are done with the same settings as CIFAR-10 while the resolution of images we use for the ImageNet test is 227x227, which is 50 times larger than CIFAR-10 (32x32). 
However, confidence results on Cifar-10 dataset is comparatively much higher 
than ImageNet. 
In each successful attack the probability label of the target class(selected by the attack) is the highest. Therefore, the average confidence on ImageNet is relatively low but tell us that the other remaining $999$ classes are even lower such that the output becomes an almost uniform soft label distribution.  To sum it up, the attack can break the confidence of AlexNet to a nearly uniform soft label distribution. 
The results indicate the large images can be less vulnerable than mid-sized images.

\begin{table}
\begin{center}
\begin{tabular}{|c|c|c|c|c|}
	\hline  Variant&Success Rate&Confidence&Cost\\
	\hline 0.5/0.5/0.5&$71.46\%$&$89.38\%$&24.66\\
	\hline 0.9/0.5/0.5&$72.00\%$&$88.22\%$&25.71\\
	\hline 0.1/0.5/0.5&$70.63\%$&$90.86\%$&20.32\\
	\hline
\end{tabular}
\end{center}
\caption{Results of conducting the proposed attack on All Convolutional network(AllConv) with different F values. Among the measures in the first row, the cost indicates the average distortion in pixel values.}
\end{table}

\begin{table}
\begin{center}
\begin{tabular}{|c|c|c|c|c|}
	\hline  Variant&Success Rate&Confidence&Cost\\
	\hline 0.5/0.5/0.5&$71.46\%$&$89.38\%$&24.66\\
	\hline 0.5/0.5/0.9&$71.66\%$&$88.60\%$&24.43\\
	\hline 0.5/0.5/0.1&$71.05\%$&$89.71\%$&24.60\\
	\hline 0.5/0.9/0.9&$72.06\%$&$90.19\%$&25.03\\
	\hline 0.5/0.9/0.5&$70.86\%$&$89.58\%$&24.69\\
	\hline 0.5/0.9/0.1&$72.06\%$&$88.70\%$&24.16\\
	\hline 0.5/0.1/0.9&$71.04\%$&$88.98\%$&24.68\\
	\hline 0.5/0.1/0.1&$72.29\%$&$88.68\%$&24.64\\
	\hline 0.5/0.1/0.5&$72.00\%$&$88.98\%$&24.86\\
	\hline
\end{tabular}
\end{center}
\caption{Results of conducting the proposed attack on All Convolutional network(AllConv) with different crossover strategies.}
\end{table}

\begin{table}
\begin{center}
\begin{tabular}{|c|c|c|c|c|}
	\hline  Variant&Success Rate&Confidence&Cost\\
	\hline
	\multicolumn{4}{|c|}{All Convolutional Net}\\
	\hline 0.1/0.1/0.1&$71.86\%$&$90.30\%$&20.44\\   
	\hline 0.5/0.1/0.1&$72.29\%$&$88.68\%$&24.64\\
	\hline
	\multicolumn{4}{|c|}{Network In Network}\\
	\hline 0.1/0.1/0.1&$77.64\%$&$99.92\%$&14.76\\   
	\hline 0.5/0.1/0.1&$78.24\%$&$99.85\%$&18.99\\
	\hline
	\multicolumn{4}{|c|}{VGG Network}\\
	\hline 0.1/0.1/0.1&$56.49\%$&$67.36\%$&22.98\\   
	\hline 0.5/0.1/0.1&$61.28\%$&$73.07\%$&24.62\\
	\hline
	\multicolumn{4}{|c|}{BVLC Network}\\
	\hline 0.1/0.1/0.1&$31.87\%$&$14.88\%$&2.36\\   
	\hline 0.5/0.1/0.1&$26.69\%$&$14.79\%$&6.19\\
	\hline
\end{tabular}
\end{center}
\caption{Results of conducting proposed attacks on additional datasets by using local-optimized DE 0.1/0.1/0.1 and 0.5/0.1/0.1.}
\end{table}




The results of attacks are competitive with previous non-targeted attack methods which need much more distortions (Table~7).

\begin{table}
\begin{center}
\begin{tabular}{|c|c|c|c|c|}
	\hline Method&\rotatebox{90}{Success rate}&\rotatebox{90}{Confidence}&\rotatebox{90}{Number of pixels}&\rotatebox{90}{Network}\\

	\hline 0.1/0.1/0.1&$77.64\%$&$99.92\%$&5($0.48\%$)&NiN\\   
	\hline 0.1/0.1/0.1&$56.49\%$&$67.36\%$&5($0.48\%$)&VGG\\   
	\hline LSA\cite{13}&$97.89\%$&$72\%$&33 ($3.24\%$)&NiN\\
	\hline LSA\cite{13}&$97.98\%$&$77\%$&30 ($2.99\%$)&VGG\\
	\hline FGSM\cite{2}&$93.67\%$&$93\%$&1024 ($100\%$)&NiN\\
	\hline FGSM\cite{2}&$90.93\%$&$90\%$&1024 ($100\%$)&VGG\\
	\hline
\end{tabular}
\end{center}
\caption{Compassion of attack effectiveness between the proposed method with DE 0.1/0.1/0.1 and two previous works, which shows that even under more restricted condition, the proposed method can still perform comparative effectiveness to previous works.}
\end{table}

\subsubsection{Original-Target Class Pairs.}
\begin{figure}[t]
\begin{center}
\includegraphics[width=0.6\linewidth]{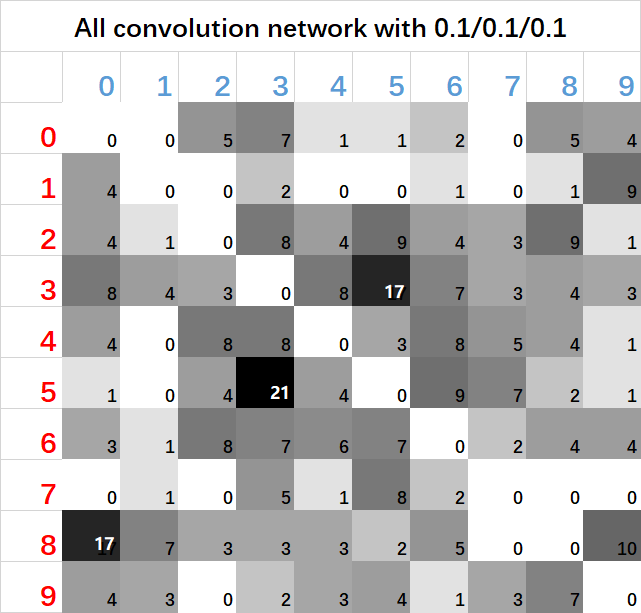}
\end{center}

\begin{center}
\includegraphics[width=0.6\linewidth]{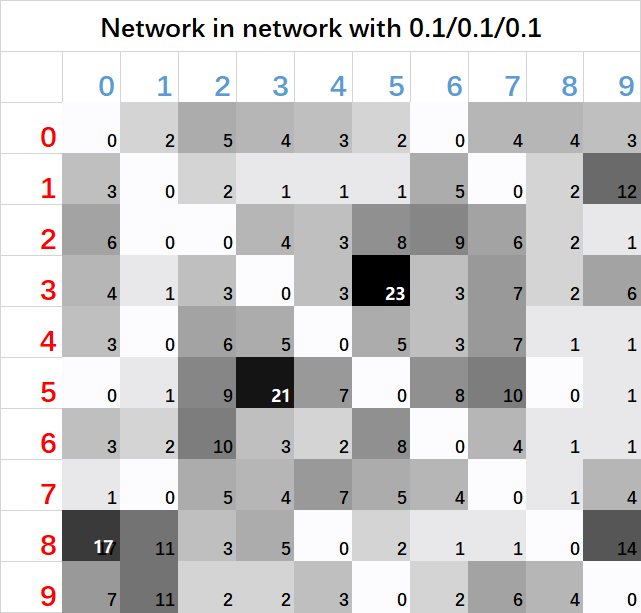}
\end{center}

\begin{center}
\includegraphics[width=0.6\linewidth]{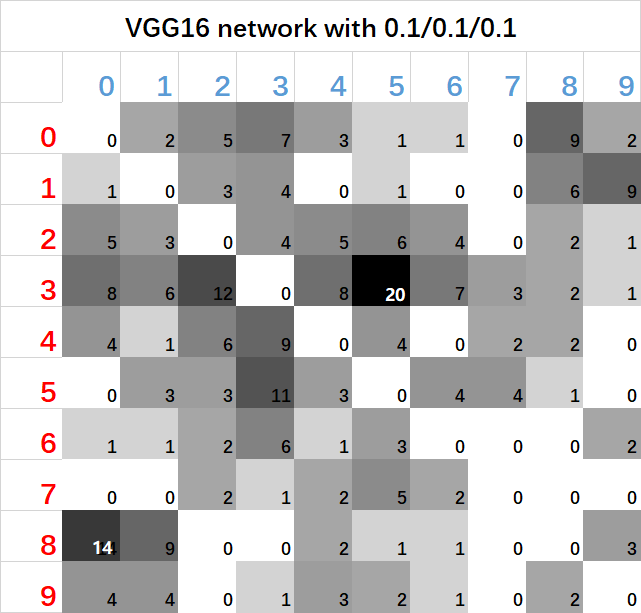}
\end{center}

\caption{Heat-maps of the number of times a successful attack is present with the corresponding original-target class pair, on three types of networks with attacks based on local-optimized DE 0.1/0.1/0.1. 
Red and blue indices indicate respectively the original and target classes. 
The number from $0$ to $9$ indicates respectively the following classes: airplane, automobile, bird, cat, deer, dog, frog, horse, ship, truck.}
\label{fig:long}
\label{fig:onecol}
\end{figure}

\begin{figure}[t]
  
\begin{center}
\includegraphics[width=0.6\linewidth]{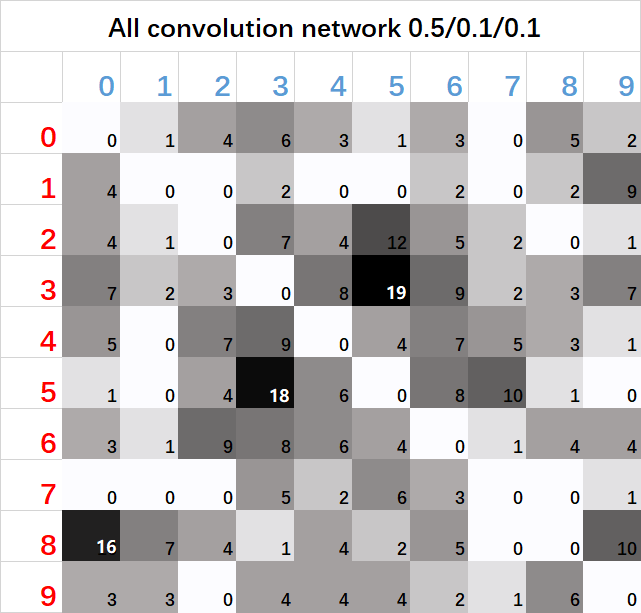}
\end{center}

\begin{center}
\includegraphics[width=0.6\linewidth]{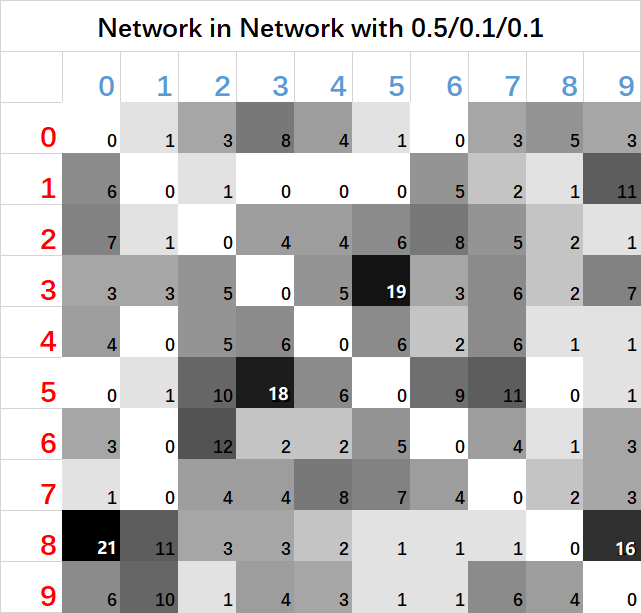}
\end{center}

\begin{center}
\includegraphics[width=0.6\linewidth]{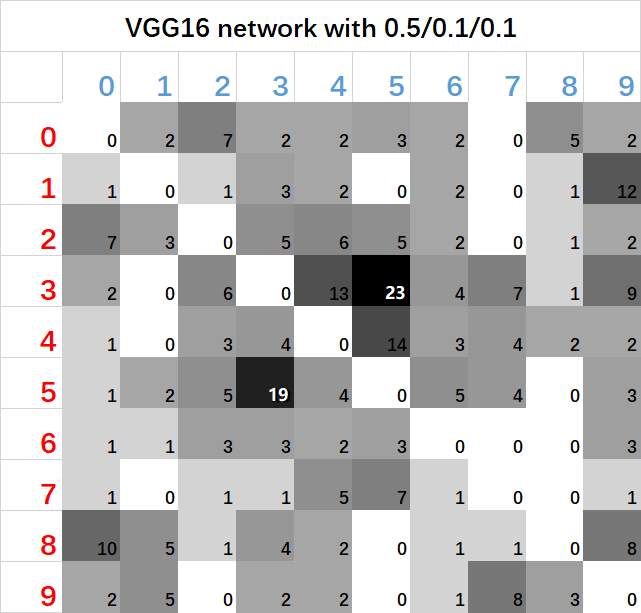}
\end{center}

\caption{Heat-maps of the number of times a successful attack is present with the corresponding original-target class pair, on three types of networks with attacks based on local-optimized DE 0.5/0.1/0.1. }
\label{fig:long}
\label{fig:onecol}
\end{figure}

\begin{figure}[t]
\begin{center}
\includegraphics[width=0.8\linewidth]{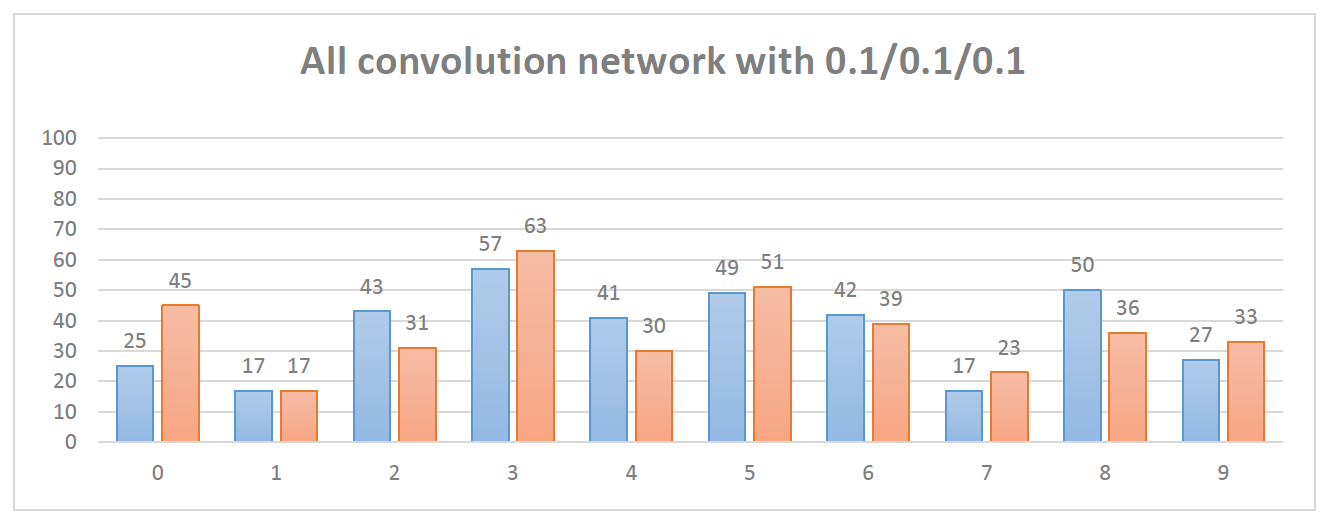}
\end{center}
\begin{center}
\includegraphics[width=0.8\linewidth]{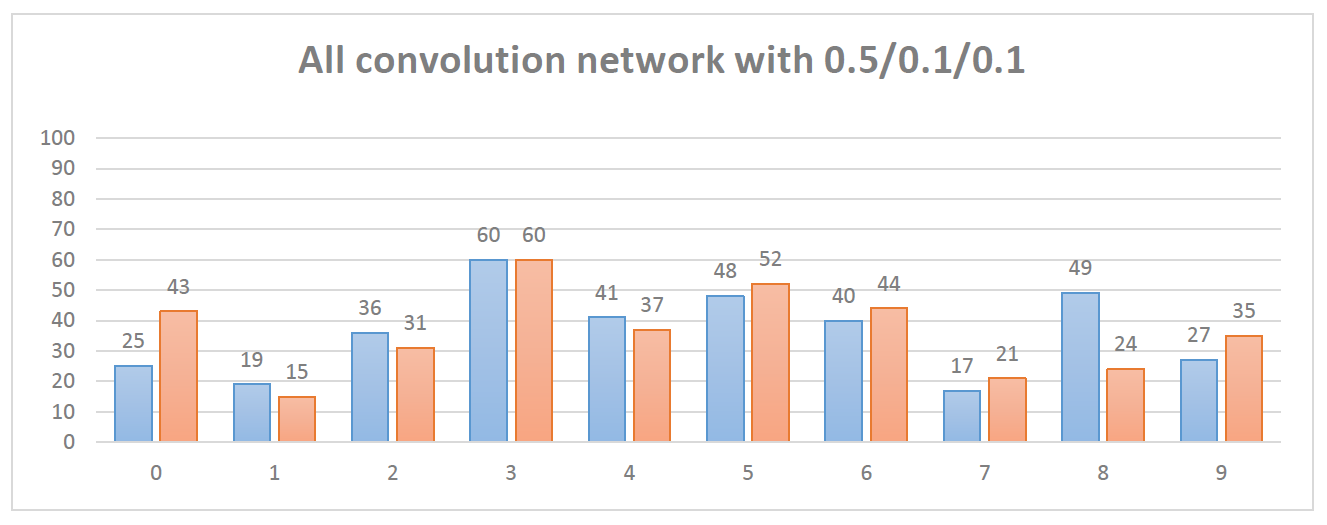}
\end{center}
\begin{center}
\includegraphics[width=0.8\linewidth]{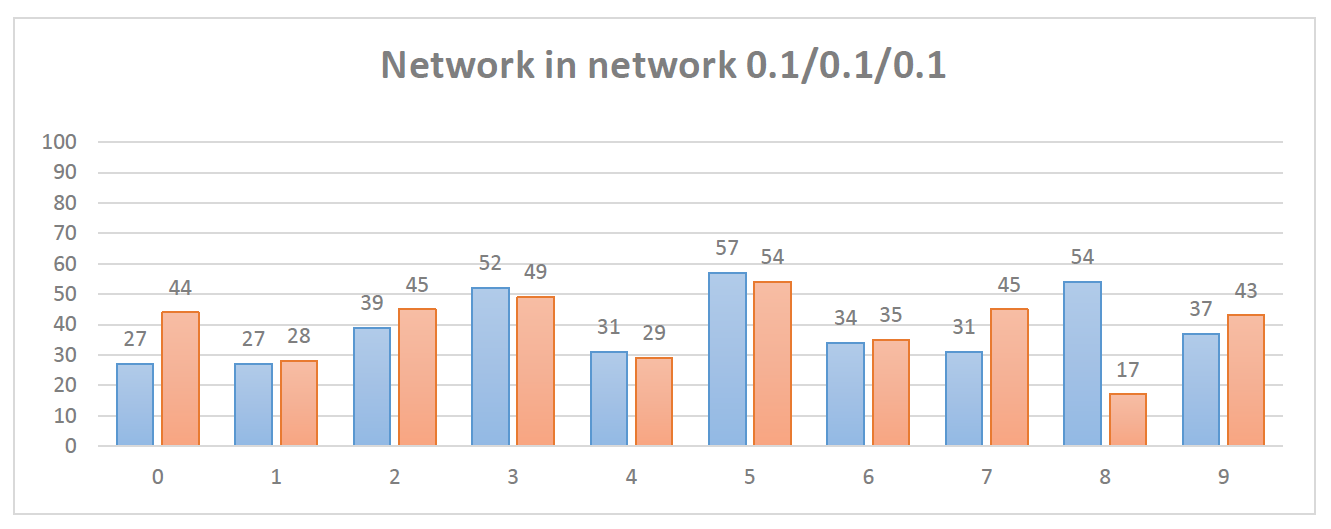}
\end{center}
\begin{center}
\includegraphics[width=0.8\linewidth]{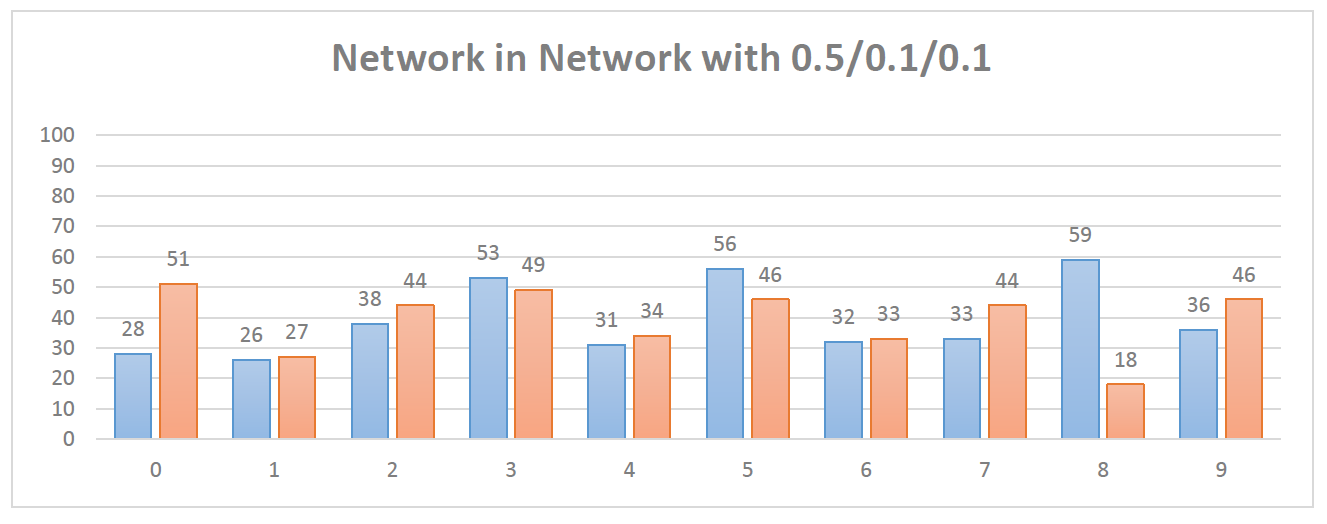}
\end{center}
\begin{center}
\includegraphics[width=0.8\linewidth]{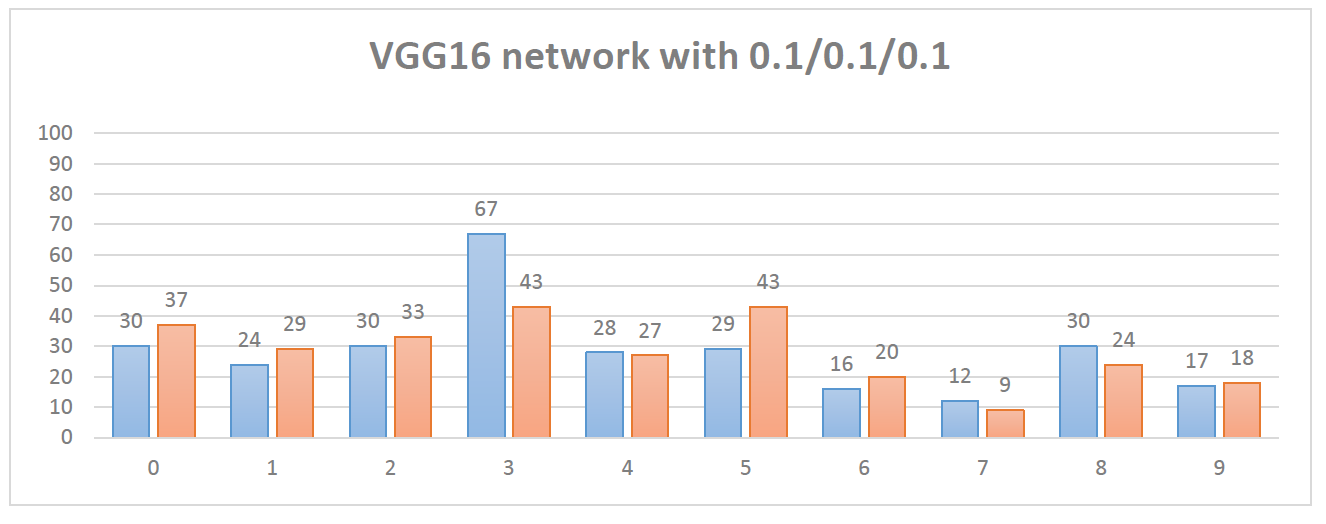}
\end{center}
\begin{center}
\includegraphics[width=0.8\linewidth]{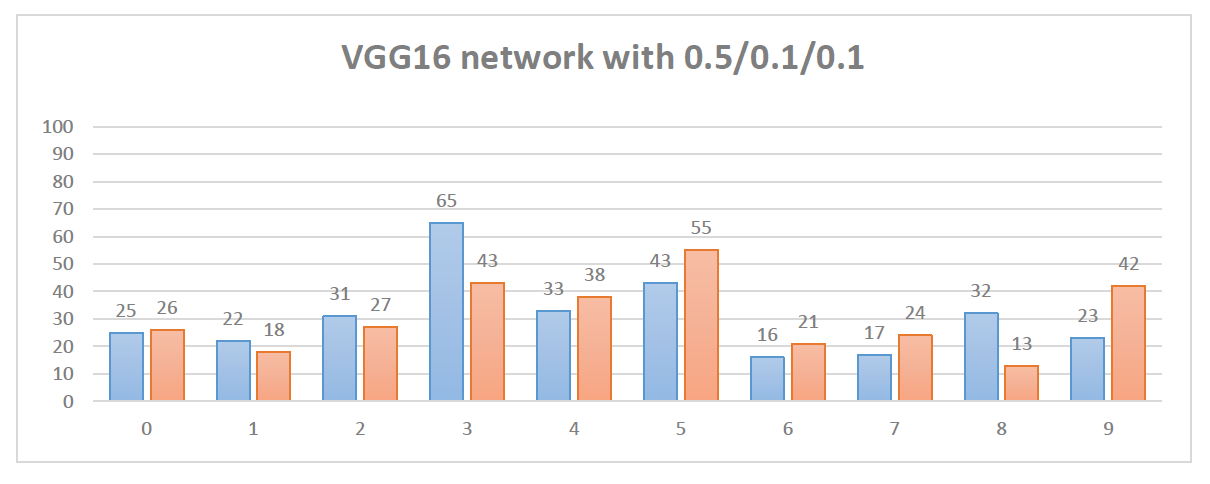}
\end{center}
   \caption{Number of successful attacks (vertical axis) for a specific class acting as the original (blue) and target (red) class. 
   The horizontal axis indicates the index of each class which is the same as Figure~5.}
\label{fig:long}
\label{fig:onecol} 
\end{figure}

Since there is no term in the fitness function used to favor the accuracy of a specific target class (i.e. non-targeted attack), the evolved perturbation is expected to trend to move the original images towards the most close target class such that the results of original-target class pairs can be seen as an indirect distance map between the original and different target classes.
For example, images of cat (class 3) is relatively much close and can be more easily perturbed to dog (class 5) through all types of networks and DE variants being tested. 

Overall, it can be seen that some certain classes can be more easily perturbed to another close target class. Even if the original and target class might be quite similar (e.g. cat and dog) for both CNN and human eyes, in practice such a vulnerability can be still fatal. In addition, the vulnerability might be even regarded as a guideline for adversaries to launch targeted attack. 
Saying that an adversary wishes a natural image with true label $C_o$ can be mis-classified to a specific target class $C_t$. According to the distance map he(she) finds that directly perturbing $C_o$ to $C_t$ is hard but it is easy to perturb $C_o$ to a third class $C_m$ which has much less distance to $C_t$. Then an option is to first perturb $C_o$ to $C_m$ and then to the final destination $C_t$. 
For example, according to the heat-map of All convolution network with 0.1/0.1/0.1 (the first graph of Figure.~4), an adversary can perturb an image with label 0 to 9 by first perturbing the image to class 8 then to class 9. Doing in such a way is easier than directly perturbing from 0 to 9.

Additionally, it can also be seen that each heat-map matrix is approximately symmetric, indicating that each class has similar number of adversarial samples which were crafted from these classes as well as to these classes, which is also directly suggested by Figure~6. There are certain classes that are apparently more vulnerable since being exploited more times than other classes, as the original and target classes of attack. The existence of such vulnerable classes can become a backdoor for inducing security problems.

\subsubsection{Time complexity}

The time complexity of DE can be evaluated according to the number of evaluations which is a common metric in optimization. Specifically, the number of evaluations is equal to the population size multiplied by the number of generations. In this research we set the maximum number of generation as 100 and population size as 400 therefore the maximum number of evaluations is 40000. We observed that all DE variants reach the maximum number of evaluations for each experiment on average. 
Even so, according to the results mentioned above, the proposed attack can produce effective solutions in such a small number of evaluations.

\section{Discussion and Future Work}

Our results show the influence of adjusting parameters of DE to the effectiveness of attack. According to the comparison between different DE variants, it can be seen that a small F value can induce little reduction on success rate of attack but reduce about $26\%$ distortion needed for conducting the attack. In practice, adversaries can choose to emphasize either success rate or distortion by adjusting the F value. The crossovers between coordinates and RGB values of the perturbation are shown to be not useful for generating better quality perturbation. Such a phenomenon can be easily realized by comparing the results between the DE that disables both crossovers and others. This might indicate that for a specific effective perturbation $x_i$, its coordinate and RGB value are strongly related. Transplanting either the isolated vulnerable coordinate or RGB value of $x_i$ to another perturbation can be no helpful, or even decrease the quality of latter. Furthermore, the result might indicate that for a specific natural image, universal vulnerable pixels or RGB values can hardly exist in contrast to the exsitence of the universal perturbation with respect to multiple images \cite{29}.
By vulnerable pixel we mean a specific pixel can be vulnerable with multiple RGB values. And vulnerable RGB value is a specific value that keeps its vulnerability across different positions on an image. In other words, our results shows that a success adversarial perturbation has to be conducted at a specific locale on the image also with a specific RGB value.
  
We show that DE can generate high quality solution of perturbation by considering realistic constraints into the fitness function. Specifically, the research evaluates the effectiveness of using DEs for producing adversarial perturbation under different parameter settings.
In addition, the DEs implemented are with low number of iterations and a relatively small set of initial candidate solutions. Therefore, the perturbation success rates should improve further by having either more iterations or a bigger set of initial candidate solutions. 

The ultimate goal of proposing attack against the CNN is evaluating its vulnerability. The CNN has been shown to have different levels of vulnerabilities to additive perturbation created from different types of optimization methods. The proposed attacks shows that CNN is even vulnerable to such a low cost, low dimensional imperceptible attack even under extremely limited conditions. The future extension can be done by analyzing and explaining why CNN is vulnerable to such various types of attacks simultaneously and according extracting possible countermeasures.

\section{Acknowledgment}
This research was partially supported by Collaboration Hubs for International Program (CHIRP) of SICORP, Japan Science and Technology Agency (JST).

{\small
\bibliographystyle{ieee}

}

\end{document}